# Machine Learning Fairness in House Price Prediction: A Case Study of America's Expanding Metropolises

ABDALWAHAB ALMAJED, Imam Abdulrahman Bin Faisal University, Saudi Arabia

MARYAM TABAR, The University of Texas at San Antonio, USA

PEYMAN NAJAFIRAD, The University of Texas at San Antonio, USA

As a basic human need, housing plays a key role in enhancing health, well-being, and educational outcome in society, and the housing market is a major factor for promoting quality of life and ensuring social equity. To improve the housing conditions, there has been extensive research on building Machine Learning (ML)-driven house price prediction solutions to accurately forecast the future conditions, and help inform actions and policies in the field. In spite of their success in developing high-accuracy models, there is a gap in our understanding of the extent to which various ML-driven house price prediction approaches show ethnic and/or racial bias, which in turn is essential for the responsible use of ML, and ensuring that the ML-driven solutions do not exacerbate inequity. To fill this gap, this paper develops several ML models from a combination of structural and neighborhood-level attributes, and conducts comprehensive assessments on the fairness of ML models under various definitions of privileged groups. As a result, it finds that the ML-driven house price prediction models show various levels of bias towards protected attributes (i.e., race and ethnicity in this study). Then, it investigates the performance of different bias mitigation solutions, and the experimental results show their various levels of effectiveness on different ML-driven methods. However, in general, the in-processing bias mitigation approach tends to be more effective than the pre-processing one in this problem domain. Our code is available at https://github.com/wahab1412/housing_fairness.

## 1 Introduction

Housing is a basic human necessity with significant impacts on individual's health, well-being, and social welfare [37]. In particular, research found that housing instability and insecurity could result in mental health issues [9], and adverse educational outcome [7]. Furthermore, inequity in housing is a challenging problem in various regions across the United States with disproportionate adverse effects on certain minority groups [12]. Therefore, improving the housing conditions for all could have significant positive social impacts.

Authors' Contact Information: Abdalwahab Almajed, Imam Abdulrahman Bin Faisal University, Dammam, , Saudi Arabia, analmajed@iau.edu.sa; Maryam Tabar, The University of Texas at San Antonio, San Antonio, Texas, USA, maryam.tabar@utsa.edu; Peyman Najafirad, The University of Texas at San Antonio, San Antonio, Texas, USA, peyman.najafirad@utsa.edu.

To improve the housing conditions, much research has been conducted in various disciplines (such as real estate, urban planning, and computer science) [23]. In particular, several studies focused on developing house price prediction models to accurately forecast the future price with the goal of informing actions and policies in the field. Especially, recent research relied on ML-driven algorithms, such as Random Forest (RF) [36, 45], XGBoost [45], and deep learning [23, 42], to accurately forecast the house price from heterogeneous sources of data (such as structural, temporal, or neighborhood-level data [23]). As a result, they found ML-driven approaches to be more effective for house price prediction, as compared to some traditional approaches. However, they mainly focused on increasing the predictive accuracy (mainly through refining input features and/or ML architecture), rather than the fairness of those solutions. Recently, a few studies [33, 48] investigate the bias of Zestimate [49], however, there is major gap in our understanding of the fairness of various other ML-driven house price prediction approaches (proposed in the past literature) and the effectiveness of various bias mitigation algorithms in this particular problem domain.

To fill this gap, this paper aims to study the following two questions:

Q1. Would various ML-driven house price prediction models show racial and ethnic bias?

Q2. If yes, how effective are bias mitigation algorithms in reducing bias in ML-driven house price prediction models?

To answer these questions, we collect comprehensive neighborhood-level features as well as structural housing attributes for 232,057 houses across San Antonio[1] in Texas. Then, to address Q1, we develop several predictive ML models commonly used for house price prediction in the past literature, and evaluate their fairness using two group-based fairness metrics, namely Independence and Separation [10]. As a result, we find that, although the ground-truth labels do not show substantial level of bias in our dataset, the models' outputs shows varying levels of bias towards protected attributes (i.e., race and ethnicity in this study).

Next, to address Q2, we build pre-processing and in-processing bias mitigation algorithms, i.e., Correlation Remover (CR) [5] and Reduction-based algorithm [1] (respectively), and investigate their effectiveness in reducing bias in this problem domain. The empirical results suggests that the effectiveness of these methods could vary across various ML models, however, in general, the Reduction-based algorithm tend to be more effective than CR in improving the fairness. In particular, on average, employing the Reduction-based algorithm results in about 23% and 6% improvement in the median of obtained Independence and Separation (respectively), while the use of CR overall does not lead to a significant improvement in our experiments.

Understanding and mitigating bias in ML-driven house price prediction models could have significant real-world implications. In particular, once an fair high-accuracy predictive tool is developed, it could have a wide range of applications in real estate market and urban planning with diverse beneficiaries (including home buyers, sellers, policymakers, etc.) [14]. For instance, these models can improve appraisal accuracy by mitigating the undervaluation of properties in minority communities [28]. Moreover, it could help policymakers make informed plans to increase homeownerships and better enhance housing stability and security in order to make cities and human settlements sustainable. Additionally, developing fair ML model aligns well with recent national efforts on promoting fair housing for all (such as the Interagency Task Force on Property Appraisal and Valuation Equity initiative [39]).

## 2 Related Work

The topic of house price valuation has been studied from different perspectives in various disciplines, including computer science, urban planning, and real estate [23]. In particular, some research (such as [41]) relied on predictive analytics

---

[1]San Antonio, the seventh most populous U.S. city, faces significant challenges in housing instability and racial homeownership gaps [46].

Manuscript submitted to ACM

(especially classical ML and deep learning models) to accurately forecast the future house price and their empirical results confirmed the value of ML-driven solution for house price prediction; for example, prior work [41] developed a combination of classical ML models (RF [6] and XGboost [11]) as well as traditional approaches (such as Hedonic regression [36]), and showed that certain ensemble decision tree-based models outperformed traditional approaches. Further, recent research [29] showed the value of Long Short-Term Memory (LSTM) networks [27] in accurately predicting house prices from time-series data (compared to classical ML models). Moreover, Riveros et al. [42] proposed a transformer-based graph model to predict house prices by leveraging the dependencies of geospatial interactions between properties, resulting in better performance compared to other well-known models such as XGBoost and RF. In addition to investigating the value of various ML architecture, past literature has studied the value of a diverse set of features, such as in-door/out-door imagery data [32], neighborhood characteristics (e.g., income [29], educational attainment [25], transportation profile [18, 29]) and structural features (e.g., number of bedroom, area, age, etc.) [23], for enhancing house price prediction. As a result, they found that certain neighborhood characteristics (such as income level) and structural features had significant contributions in the predictive performance of their house price prediction approaches [41]. Although these findings are useful for understanding the value of ML and various factors for house price prediction, they mainly focused on increasing the predictive accuracy through enhancing input features and/or ML architecture and did not study if those ML models show racial and ethnic bias. There have been a few studies [33, 48] regarding the bias of one commercially available house price prediction model, namely Zestimate [49], and found that its level of bias is relatively small [33] and leveraging Zestimate could potentially reduce racial bias in this specific application domain [48]. However, they only studied a single model and there is still a gap in our understanding of the level of bias of various ML-driven house price prediction approaches (proposed in past literature) under different privileged group definitions based on racial and/or ethnic composition of neighborhood residents, and also, how effective the bias mitigation algorithms are in reducing bias across various models in this problem domain.

Some other research has relied on descriptive analytics to study the variations in housing price across various neighborhoods with different racial and ethnicity composition. As a result, they found disparity in housing price and its growth in privileged vs unprivileged neighborhoods. For example, past literature [31] found that residences in Black and Hispanic communities are typically located in areas more prone to substantial value depreciation arising from distressed sales. Additionally, previous work [21] examined the impact of race on house values during a housing crisis, revealing that homes in predominantly Black neighborhoods witnessed more significant declines in value and a slower recovery compared to those in predominantly White neighborhoods. Moreover, research [26] revealed that Black individuals face greater challenges in accessing housing in high-opportunity areas due to financial and spatial constraints, limiting their opportunities for wealth accumulation. Although their results are helpful for understanding disparities in the housing market, they did not study the algorithmic fairness of ML-driven house price prediction solutions and mitigation of the potential biases, which will be studied here.

## 3 Datasets

Our study mainly relies on the following publicly-available sources of data:

(1) **BexarCAD Data [19]:** This data consists of value of houses apprised and released by Bexar Appraisal District (BexarCAD). The website provides access to property valuations, and we collect annual house prices from 2020 to 2023. BexarCAD appraises home values based on fair market value –the price a property would sell for under current market conditions. In other words, the appraised values reflect the true value of homes in today's market.

  (2) **American Community Survey (ACS) Data [8]:** This dataset includes annual data on socio-economic characteristics of various census tracts across the United States. Inspired by the findings of prior work on the strong association of neighborhood characteristics with house price [23, 41], we incorporate such data into the prediction process. In particular, we extract the following characteristics: financial factors (such as mean income, percentage of population below poverty line), employment status of residents (along with the number of people working in various industries), educational attainment, race/ethnicity composition, housing-related attributes (such as the percentage of houses across various price ranges and the percentage of renter-occupied housing), and the residents' means of transportation to work. For our prediction task, we incorporate time-series ACS data of the preceding $k$ time steps available ($k$ is set to 9) as the input to capture changes in the neighborhood characteristics over time and model their association with house price effectively. Please note race/ethnicity is not fed as input to ML models, and is only used for bias measurement/mitigation.
  (3) **Redfin Data [16]:** The Redfin website (www.redfin.com) releases house-specific data for properties across the United States, and we use this website to crawl several structural characteristics of houses, such as the number of bedrooms and the total areas in sqft, which in turn, have been found to impact the house price [23]. This data provides static input features for our prediction task.

**Data Pre-Processing Steps.** Our dataset comprehensively encompasses data on 232,057 residential properties. Each property is characterized by 5 static attributes, in addition to 52 time-varying input features that evolve over time. We selected five static features that represent the physical characteristics of the houses, which prior research [23] has shown to be effective for accurate house price prediction. Additionally, we included dynamic features that capture demographic changes in the neighborhood over time, as such shifts can influence how property values evolve in certain areas. Data pre-processing begins with preparing the target variable, framing the problem as a next-step prediction task (regression). Following prior research [29], first, we subtract the house price at time $t-1$ (i.e., $y_{t-1}$) from house price at time $t$ (i.e., $y_t$). However, we note that this could be biased towards protected groups because, per previous research [21], house prices in privileged neighborhoods could be significantly higher than those in unprivileged neighborhoods. This natural bias in the data could normally make the ML models biased as well. To address this, we consider the percentage of house price increases (i.e., $(y_t - y_{t-1})/y_{t-1}$) as our target variable, which does not show a tangible dependence to the protected characteristics in our data. Next, we considered the data of 2021, 2022 and 2023 as our training, validation, and test sets, respectively, and also, conduct min-max normalization.

## 4  Q1: Fairness Assessment in ML-Driven House Price Prediction Models

Our first research question aims to study the racial and ethnic bias of various ML-driven house price prediction models under different definitions of privileged groups. In fact, prior work investigated the importance of various sources of data (such as structural, imagery, description, etc.) and ML models for house price prediction, and as a result, they found that neighborhood-related features (such as income) are often among the most important set of features [23, 29]. However, there is still a gap in understanding of the fairness of various ML-driven house price prediction models under various definitions of privileged groups and the correlation of certain neighborhood characteristics with protected attributes [24] motivated us to study the extent to which various ML-driven approaches relying on such input features show racial/ethnic bias in this problem domain.

In particular, we assess the fairness using two popular group-based fairness metrics: (1) Independence [10], and (2) Separation [10]. Independence assesses how much the model's predictions are dependent to the protected attribute, and Separation measures the level of conditional independence between a model's predictions and the protected attribute, given the ground-truth value [10]. While these measures can be easily computed in the classification domain with categorical labels, they are not easy to compute in the regression setting with continuous labels [43]. To address this issue, several methods (such as [43]) were introduced to estimate (rather than exactly measure) these fairness metrics in the regression domain. Following the same approach as in [2], we incorporate four algorithms, including P1 [1], P2 [15], P3 [34] and P4 [4, 43], to measure Independence and we leverage C1 [4, 43] and C2 [34] to measure Separation. Please note that, for P1, P2, and P3, and C2, the level of deviation from 0.0 shows the level of bias, while for P4 and C1, the amount of deviation from 1.0 shows the level of bias.

To comprehensively study fairness in this problem domain, we measure the fairness metrics under three definitions of the privileged group, which rely on race, ethnicity, or both:

(1) The first definition relies only on the racial composition of a neighborhood, and accordingly, houses in white-majority census tracts[2] fall under the privileged group in this case.
(2) The second definition relies only on ethnicity, and accordingly, houses in census tracts with majority of non-Hispanic belong to the privileged group in this case.
(3) The third definition relies on both race and ethnicity, and accordingly, houses in census tracts with the majority of Non-Hispanic White population fall under the privileged group in this case.

Table 1 shows the total number of houses in privileged and unprivileged groups based on the criteria used for defining privileged groups.

| Criteria | Privileged | Unprivileged |
| --- | --- | --- |
| Race | 226,164 | 5,893 |
| Ethnicity | 70,526 | 161,531 |
| Race & Ethnicity | 66,693 | 165,364 |

Table 1. Number of houses in privileged/unprivileged neighborhoods based on the criteria defining privileged groups.

As for ML models, we consider a collection of classical ML models and neural networks, specifically XGBoost [11], RF [6], Multi-Layer Perceptron (MLP), TabNet [3], LSTM [27], and Gated Recurrent Units (GRU) [13]. We select these models because of several reasons. First, many prior studies in house price prediction relied on XGboost and RF, and achieved a high predictive performance [30, 36, 45]. Additionally, recent research showed that ensemble decision-tree based models outperform various state-of-the-art time-series prediction models in several application domains in terms of the predictive performance [22]. We also considered TabNet (which is an attention-based neural network model) due to its potential in outperforming decision tree-based ensemble models [3]. Additionally, LSTM and GRU have shown promising results in various time-series prediction tasks and especially, house price prediction [29], which in turn, motivated our selection of these algorithms.

**Experimental Results.** With this background, we now provide the results of our experiments. Table 2 shows the predictive performance of the aforementioned ML models and the baseline in terms of Root Mean Square Error (RMSE). The first value (LastYear) shows a baseline that returns the exact percentage of change in house price for the preceding year as the the predicted value (and so, does not have any ML component), and then, other values correspond to the performance of our ML-driven models. According to the results, all models outperform the baseline, which confirms the value of those ML models along with our input variables for this task. Also, LSTM and GRU outperform all models, particularly they achieve 55.1% lower RMSE than the baseline (last year). Having established the effectiveness of our ML models, we now evaluate the fairness of those ML models.

---

[2]Please note that, although race has more than two categories, the neighborhoods in our dataset are either white-majority or black-majority.

| Model | LastYear | XGBoost | RF | MLP | TabNet | LSTM | GRU |
|---|---|---|---|---|---|---|---|
| **RMSE** | 0.165 | 0.075 | 0.077 | 0.079 | 0.082 | 0.074 | 0.074 |

Table 2. Predictive performance of house price prediction models in terms of RMSE.

To this end, first, we investigate the level of bias in our test set (to be able to consider its leakage into the overall bias evaluation). Table 3 shows the level of bias in our test set, which suggests no substantial bias in most cases.

| Criteria | P1 | P2 | P3 | P4 |
|---|---|---|---|---|
| Race | 0.19 | 0.19 | 0.00 | 1.07 |
| Ethnicity | 0.02 | 0.03 | 0.07 | 1.00 |
| Race & Ethnicity | 0.03 | 0.05 | 0.08 | 1.00 |

Table 3. The level of bias in the house price in the test set under various criteria used for defining the privileged group.

Next, we measure bias in ML models' predictions. Table 4 shows the results of fairness evaluation based on our three privileged group definitions using various fairness measurement methods. According to the results, various models show varying levels of bias, with the bias level changing based on the criteria for defining privileged groups. Notably, RF, a commonly used model in house price prediction, exhibits the highest bias across all metrics when both race and ethnicity are used to define the privileged group.

**Feature Importance Analysis.** While RMSE and the fairness metrics provide objective measures on the performance of the ML models, it is important to dig deeper to gain further insights into the behavior of models, and investigate what features seem to contribute the most in various circumstances. To this end, we conduct a feature importance analysis on XGBoost model, which shows high level of bias in its outcome, and is more explainable by nature due to its mechanics. Table 5 shows the top-5 most important features for XGBoost along with their importance values. Based on the results, XGBoost tends to highly rely certain neighborhood characteristics, which have been found to be strongly correlated with protected attributes. In particular, we see that a few employment-related factors (including the unemployment rate) show up among the top-5 most important features, and prior work [38] found gaps in unemployment rate across various racial groups. Additionally, the percentage of renter-occupied housing units is among the top-5 most important features, and prior research found racial gap in homeownership [46], and considerable racial and ethnic disparities among who rents and who owns the homes [20]; e.g., while over half of the African-American or Latino-led households live in rental houses, about 27% of households headed by Non-Hispanic White is reportedly rental [20]. Therefore, it is not unexpected to see bias in the performance of an ML model heavily relying on such neighborhood characteristics that tend to be strongly correlated with race and/or ethnicity. With that, next, we will focus on the effectiveness of bias reduction algorithms in mitigating the models' bias towards protected attributes.

| Criteria | Model | P1 | P2 | P3 | P4 | C1 | C2 |
|---|---|---|---|---|---|---|---|
| Race | XGBoost | 0.30 | 0.31 | 0.00 | 1.51 | 1.45 | 0.00 |
| | RF | 0.30 | 0.31 | 0.21 | 1.33 | 1.29 | 0.19 |
| | MLP | 0.24 | 0.24 | 0.00 | 1.09 | 1.07 | 0.00 |
| | TabNet | 0.29 | 0.30 | 0.00 | 1.21 | 1.17 | 0.14 |
| | LSTM | 0.16 | 0.17 | 0.00 | 1.06 | 1.06 | 0.00 |
| | GRU | 0.25 | 0.25 | 0.00 | 1.12 | 1.12 | 0.00 |
| Ethnicity | XGBoost | 0.25 | 0.37 | 0.36 | 1.28 | 1.29 | 0.51 |
| | RF | 0.37 | 0.53 | 0.53 | 1.65 | 1.72 | 0.64 |
| | MLP | 0.30 | 0.43 | 0.49 | 1.41 | 1.43 | 0.60 |
| | TabNet | 0.18 | 0.26 | 0.29 | 1.09 | 1.10 | 0.42 |
| | LSTM | 0.06 | 0.09 | 0.15 | 1.00 | 1.00 | 0.30 |
| | GRU | 0.09 | 0.13 | 0.17 | 1.04 | 1.04 | 0.28 |
| Race & Ethnicity | XGBoost | 0.28 | 0.39 | 0.38 | 1.34 | 1.35 | 0.52 |
| | RF | 0.40 | 0.57 | 0.56 | 1.81 | 1.89 | 0.65 |
| | MLP | 0.31 | 0.44 | 0.50 | 1.45 | 1.47 | 0.61 |
| | TabNet | 0.20 | 0.29 | 0.31 | 1.12 | 1.12 | 0.43 |
| | LSTM | 0.07 | 0.10 | 0.16 | 1.00 | 1.00 | 0.30 |
| | GRU | 0.11 | 0.15 | 0.19 | 1.04 | 1.04 | 0.28 |
| Average | - | 0.23 | 0.30 | 0.24 | 1.25 | 1.26 | 0.33 |
| Median | - | 0.25 | 0.30 | 0.20 | 1.17 | 1.15 | 0.30 |

Table 4. Results of fairness assessment of ML-driven house price prediction models under various criteria for defining the privileged group (i.e., race, ethnicity, or both).

| Feature Description | Imp. |
|---|---|
| Unemployment rate | 0.026 |
| % of housing of value between $35K and $40K | 0.016 |
| % of housing of value between $10K and $15K | 0.015 |
| % of civilians employed | 0.014 |
| % of renter-occupied housing | 0.013 |

Table 5. The top 5 most important features for XGBoost in the house price prediction task.

## 5 Q2: Effectiveness of Bias Mitigation Methods

Given the existing bias in the outcome of house price prediction models (per the findings from Q1), it is essential to address these biases in order to promote fairness. This section studies the effectiveness of bias mitigation algorithms proposed for regression tasks in this problem domain, and in particular, it explores two bias mitigation directions, pre-processing and in-processing methods [10]. The mitigation through pre-processing occurs prior to the training of ML models, with a primary focus on the data itself rather than the training process or the model's outcomes [10]. On the other hand, in-processing approaches mainly focus on incorporating the fairness constraints into the training process, for example, through developing new loss functions [10]. In this work, we study two bias mitigation techniques: the Correlation Remover algorithm (a pre-processing approach) [5] and Reduction-based algorithm (an in-processing approach) [1].

**(1) Correlation Remover (CR)**: This algorithm [5] is designed to transform input features in the training data to reduce biases. In fact, instances may arise where sensitive data show an apparent correlation with non-sensitive data; for instance, there might be a correlation between membership in a minority group and acceptance rates, indicating a lower likelihood of acceptance compared to other groups. This correlation, even after removing sensitive data, acts as a proxy for sensitive data [10]. CR first conducts correlation analysis between sensitive and non-sensitive data, and evaluate the extent of interdependence. Upon identifying the correlation, the algorithm conducts a linear transformation on the input features to decrease the degree of correlation in the training set while preserving the underlying meaning of the original data [5]. In our study, using the FairLearn package [5, 10], we apply a separate CR for each of the three protected groups, which leads to a new training set for each definition of protected group. In the scenario where only race is used as the protected attribute, the analysis revealed that the foremost five correlated features pertained to ACS data on the house value distribution in neighborhood, occupation, and education. Regarding the other two scenarios, the primary five correlated features were related to educational attainment. Following the elimination of correlation, we train the ML models on the refined training set from scratch.

**(2) Reduction-based Algorithm:** This method [1] aims to address bias through introducing an statistical parity-based loss term to ensure that a model functions consistent and impartial across different groups, thus avoiding systematic discrepancies in predictions based on protected attributes. It, first, conducts discretization on the target variable, and then, converts the fair regression task into the task of classification under fairness constraints [1].

Tables 6 and 7 represent the fairness metrics and RMSE of the ML-driven house prediction models after applying CR and Reduction-based algorithm (respectively) under various privileged group definitions. According to the results, in general, the Reduction-based algorithm tends to be more effective than CR in improving the fairness (while not significantly affecting RMSE) across various ML models and fairness metrics. In particular, on average, employing the Reduction-based algorithm improved the median [3] of values achieved for Independence and Separation by about 23% and 6% respectively (while employing CR does not result in tangible improvement overall). However, we note some exceptions as well; for example, both CR and Reduction-based algorithms seem to consistently damage the fairness of TabNet outcome when ethnicity is involved in the definition of privileged group.

One potential reason behind the different performance of the two methods could be related to their underlying assumptions and procedures. For example, CR mainly targets linear associations between the input features and sensitive attribute, while there might be some non–linear dependencies that cannot be addressed by CR [5]. In contrast, Reduction–based algorithm attempts to address bias during optimization and iteratively balance prediction error against fairness constraint violations [1]. This way the model can adapt its parameters until it reaches a certain balance of accuracy and the fairness level. Therefore, the level of effectiveness of bias reduction algorithms could vary significantly, depending on the nature of data, ML model, and the definition of privileged group.

In summary, these results collectively suggest that the bias mitigation algorithms could have varying levels of impacts on reducing the algorithmic bias of different ML models. Further, a data pre–processing technique alone might not necessarily be enough for mitigating bias in certain situations in the house price prediction problem domain, and modifying the learning process through applying Reduction–based algorithm tends to be more effective in certain cases.

---

[3]We use the median to reduce the influence of outlier values that could skew the results.

| Criteria | Model | RMSE | P1 | P2 | P3 | P4 | C1 | C2 |
|---|---|---|---|---|---|---|---|---|
| **Race** | XGBoost | 0.075 (0.000) | 0.18 (-0.12) | 0.18 (-0.13) | 0.00 (0.00) | 1.01 (-0.50) | 1.00 (-0.45) | 0.00 (0.00) |
| | RF | 0.077 (-0.001) | 0.46 (0.16) | 0.47 (0.16) | 0.26 (0.05) | 1.02 (-0.31) | 1.01 (-0.28) | 0.25 (0.06) |
| | MLP | 0.079 (0.000) | 0.28 (0.04) | 0.29 (0.05) | 0.00 (0.00) | 1.08 (-0.01) | 1.06 (-0.01) | 0.00 (0.00) |
| | TabNet | 0.079 (-0.003) | 0.37 (0.08) | 0.38 (0.08) | 0.14 (0.14) | 1.05 (-0.16) | 1.03 (-0.14) | 0.13 (-0.01) |
| | LSTM | 0.074 (0.000) | 0.34 (0.18) | 0.35 (0.18) | 0.00 (0.00) | 1.57 (0.51) | 1.58 (0.52) | 0.00 (0.00) |
| | GRU | 0.074 (0.000) | 0.16 (-0.09) | 0.16 (-0.09) | 0.00 (0.00) | 1.01 (-0.11) | 1.01 (-0.11) | 0.00 (0.00) |
| **Ethnicity** | XGBoost | 0.075 (0.000) | 0.24 (-0.01) | 0.35 (-0.02) | 0.30 (-0.06) | 1.10 (-0.18) | 1.11 (-0.18) | 0.39 (-0.12) |
| | RF | 0.077 (-0.001) | 0.24 (-0.13) | 0.35 (-0.18) | 0.35 (-0.18) | 1.18 (-0.47) | 1.18 (-0.54) | 0.42 (-0.22) |
| | MLP | 0.079 (0.000) | 0.21 (-0.09) | 0.30 (-0.13) | 0.33 (-0.16) | 1.11 (-0.30) | 1.12 (-0.31) | 0.43 (-0.17) |
| | TabNet | 0.079 (-0.003) | 0.27 (0.09) | 0.39 (0.13) | 0.34 (0.05) | 1.18 (0.09) | 1.19 (0.09) | 0.54 (0.12) |
| | LSTM | 0.075 (0.001) | 0.06 (0.00) | 0.09 (0.00) | 0.11 (-0.04) | 1.01 (0.01) | 1.01 (0.01) | 0.31 (0.01) |
| | GRU | 0.075 (0.001) | 0.09 (0.00) | 0.13 (0.00) | 0.13 (-0.04) | 1.02 (-0.02) | 1.02 (-0.02) | 0.22 (-0.06) |
| **Race & Ethnicity** | XGBoost | 0.075 (0.00) | 0.32 (0.04) | 0.44 (0.05) | 0.42 (0.04) | 1.41 (0.07) | 1.41 (0.06) | 0.52 (0.00) |
| | RF | 0.077 (-0.001) | 0.51 (0.11) | 0.72 (0.15) | 0.77 (0.21) | 4.05 (2.24) | 4.18 (2.29) | 0.83 (0.18) |
| | MLP | 0.079 (0.000) | 0.32 (0.01) | 0.45 (0.01) | 0.46 (-0.04) | 1.47 (0.02) | 1.50 (0.03) | 0.56 (-0.05) |
| | TabNet | 0.079 (-0.003) | 0.24 (0.04) | 0.34 (0.05) | 0.38 (0.07) | 1.17 (0.05) | 1.18 (0.06) | 0.46 (0.03) |
| | LSTM | 0.074 (0.000) | 0.04 (-0.03) | 0.06 (-0.04) | 0.13 (-0.03) | 1.00 (0.00) | 1.00 (0.00) | 0.30 (0.00) |
| | GRU | 0.075 (0.001) | 0.05 (-0.06) | 0.06 (-0.09) | 0.10 (-0.09) | 1.00 (-0.04) | 1.00 (-0.04) | 0.27 (-0.01) |
| **Average** | - | 0.076 (-0.001) | 0.24 (0.01) | 0.31 (0.01) | 0.23 (-0.01) | 1.30 (0.05) | 1.31 (0.05) | 0.31 (-0.02) |
| **Median** | - | 0.076 (-0.001) | 0.24 (-0.01) | 0.35 (0.05) | 0.20 (0.00) | 1.09 (-0.08) | 1.09 (-0.06) | 0.31 (0.01) |

Table 6. Fairness measures and RMSE of ML-driven house price prediction models after applying the CR algorithm with various criteria for defining the privileged group (i.e., race, ethnicity, or both). The values in parenthesis show the amount of change in the fairness metrics and RMSE after applying the bias mitigation algorithm.

## 6 Potential Real-World Impacts

Such a predictive approach could be used by a wide range of stakeholders, ranging from government/non-government organizations to individuals in the housing-related markets. In fact, the real estate sector plays a key role in both the global and local economies, involving various stakeholders such as investors, developers, realtors, and customers. The primary driving force in this economic landscape is house prices, making accurate prediction of these prices essential for all involved parties. Real estate stakeholders rely on automated house valuation models (AVMs) in their efforts to estimate house values [14]. For example, investors and developers utilize these predictions to estimate expected profits and manage costs efficiently [18]. Moreover, precise predictions of property prices provide significant advantages to homeowners and customers who are informed about housing values while interacting with other parties [18]; for example, it could help determine the best time to buy or sell the houses and the potential amount of profit/loss over time. Biased performance of such house price prediction models could have significant adverse impacts on homeowners in unprivileged neighborhoods, where their homes may be undervalued which can effect their ability to sell their properties at a fair market price. Additionally, prospective buyers may be adversely affected by receiving financing offers that are insufficient to cover the actual cost of homes, thereby limiting their access to homeownership.

Moreover, accurate and fair ML-driven house price prediction models can also help mitigate the issue of loan denial, which is often linked to property underappraisal. Home purchases commonly rely on loans, but one major reason for loan rejection is an appraisal that falls below the contract price [28]. A report [28] found that around 15% of appraisals in Hispanic communities and 12.5% in Black communities were below the contract price, compared to just 7.4% in White communities. Additionally, Black applicants were found to be more likely to face loan denials than White applicants for underappraisal reasons [28], further widening the homeownership gap. To mitigate this issue, fair ML-driven models with a high predictive accuracy can provide unbiased and data-driven price values, ensuring valuations are accurate, up-to-date, and free from the biases present in traditional appraisal practices.

| Criteria | Model | RMSE | P1 | P2 | P3 | P4 | C1 | C2 |
| --- | --- | --- | --- | --- | --- | --- | --- | --- |
| **Race** | XGBoost | 0.075 (0.000) | 0.28 (-0.02) | 0.28 (-0.03) | 0.00 (0.00) | 1.20 (-0.31) | 1.15 (-0.30) | 0.00 (0.00) |
| | RF | 0.074 (-0.004) | 0.32 (0.02) | 0.33 (0.02) | 0.21 (0.00) | 1.28 (-0.05) | 1.23 (-0.06) | 0.22 (0.03) |
| | MLP | 0.073 (-0.006) | 0.18 (-0.06) | 0.19 (-0.05) | 0.00 (0.00) | 1.15 (0.06) | 1.11 (0.04) | 0.03 (0.03) |
| | TabNet | 0.070 (-0.012) | 0.23 (-0.06) | 0.24 (-0.06) | 0.03 (0.03) | 1.02 (-0.19) | 1.00 (-0.17) | 0.08 (-0.06) |
| | LSTM | 0.087 (0.13) | 0.16 (0.00) | 0.16 (-0.01) | 0.00 (0.00) | 1.02 (-0.04) | 1.02 (-0.04) | 0.00 (0.00) |
| | GRU | 0.072 (-0.002) | 0.14 (-0.11) | 0.14 (-0.11) | 0.00 (0.00) | 1.15 (0.03) | 1.15 (0.03) | 0.01 (0.01) |
| **Ethnicity** | XGBoost | 0.074 (-0.001) | 0.09 (-0.16) | 0.14 (-0.23) | 0.16 (-0.20) | 1.04 (-0.24) | 1.04 (-0.25) | 0.33 (-0.18) |
| | RF | 0.074 (-0.004) | 0.44 (0.07) | 0.63 (0.10) | 0.71 (0.18) | 3.25 (1.60) | 3.27 (1.55) | 0.80 (0.16) |
| | MLP | 0.074 (-0.005) | 0.12 (-0.18) | 0.17 (-0.26) | 0.19 (-0.30) | 1.04 (-0.37) | 1.04 (-0.39) | 0.37 (-0.23) |
| | TabNet | 0.072 (-0.010) | 0.32 (0.14) | 0.46 (0.20) | 0.51 (0.22) | 1.26 (0.17) | 1.28 (0.18) | 0.66 (0.24) |
| | LSTM | 0.078 (0.004) | 0.09 (0.03) | 0.13 (0.04) | 0.16 (0.01) | 1.01 (0.01) | 1.01 (0.01) | 0.25 (-0.05) |
| | GRU | 0.078 (0.004) | 0.07 (-0.02) | 0.10 (-0.03) | 0.24 (0.07) | 1.00 (-0.04) | 1.00 (-0.04) | 0.32 (0.04) |
| **Race & Ethnicity** | XGBoost | 0.074 (-0.001) | 0.10 (-0.18) | 0.14 (-0.25) | 0.19 (-0.19) | 1.03 (-0.31) | 1.03 (-0.32) | 0.33 (-0.19) |
| | RF | 0.072 (-0.006) | 0.47 (0.07) | 0.66 (0.09) | 0.73 (0.17) | 3.41 (1.60) | 3.52 (1.63) | 0.79 (0.14) |
| | MLP | 0.072 (-0.007) | 0.18 (-0.13) | 0.25 (-0.23) | 0.27 (-0.23) | 1.04 (-0.41) | 1.04 (-0.43) | 0.40 (-0.21) |
| | TabNet | 0.068 (-0.014) | 0.49 (0.29) | 0.69 (0.40) | 0.71 (0.40) | 3.19 (2.07) | 3.42 (2.30) | 0.83 (0.40) |
| | LSTM | 0.075 (0.001) | 0.08 (0.01) | 0.11 (0.01) | 0.14 (-0.02) | 1.01 (0.01) | 1.01 (0.01) | 0.25 (-0.05) |
| | GRU | 0.076 (0.002) | 0.04 (-0.07) | 0.05 (-0.10) | 0.08 (-0.11) | 1.00 (-0.04) | 1.00 (-0.04) | 0.21 (-0.07) |
| **Average** | - | 0.074 (-0.003) | 0.21 (-0.02) | 0.27 (-0.03) | 0.24 (0.00) | 1.45 (0.20) | 1.46 (0.20) | 0.33 (0.00) |
| **Median** | - | 0.074 (-0.003) | 0.17 (-0.08) | 0.18 (-0.12) | 0.18 (-0.02) | 1.04 (-0.13) | 1.04 (-0.11) | 0.29 (-0.01) |

Table 7. Fairness measures and RMSE of ML-driven house price prediction models after applying the Reduction-based algorithm with various criteria for defining the privileged group (i.e., race, ethnicity, or both). The values in parenthesis show the amount of change in the fairness metrics and RMSE after applying bias mitigation algorithm.

In addition, the fair performance of ML-driven house price predictions highly aligns with the national and global movements and policies regarding equality in housing (such as Property Appraisal and Valuation Equity [39] and fair housing [47]). In fact, the aspiration for homeownership is a common goal for many individuals, as it contributes to stability and wealth accumulation over time. However, the occurrence of bias in house appraisal could have adverse effects on people's lives. Homes, as appreciating assets, hold the potential for financial security [40]. Nevertheless, undervaluation of a house can detrimentally impact homeowners seeking to sell, profit, refinance, or secure a loan for entrepreneurial endeavors [17].

Thus, efforts to mitigate biases in this context are crucial, and hence, our approach could contribute to advancing towards reducing bias through revealing the underlying bias in ML-driven housing prediction models and providing insights on the effectiveness of popular bias mitigation algorithms.

## 7 Limitations and Future Work

This study has a few limitations that are mainly related to the dataset used in the research. First, it only relies on the data of one city (i.e., San Antonio, TX), which arises a question about the generalizability of the obtained results to other regions across the United States. In fact, San Antonio is among the four cities in the United States with the largest housing shortage [35], and has a history of residential and economic segregation [44], and racial gap in homeownership [46]. Because housing-related history and racial/ethnic composition of different cities could vary significantly across the nation, it is not impossible to observe variations in the performance of house price prediction models across the nation. Thus, one potential future work is to study the generalizability of obtained results to broader geographical contexts.

Furthermore, due to the lack of individual-level data on the race/ethnicity of each household [48], the privileged/unprivileged groups are defined based on the race and/or ethnicity of all residents in each census tract; i.e., when the privileged group definition only relies on the race, if a house is located in a White-majority census tracts, that house counts as a sample of privileged data point in our evaluation, regardless of the race of its residents. Accordingly, one future research pathway could be to collect fine-grained socio-demographic information about each household, and then, investigate the extent to which the fairness results obtained from our neighborhood-level privileged group definition are in line with those achieved in the situation where household-level demographic information is used in experiments.

## 8 Conclusion

This paper assesses the fairness of ML-driven house price prediction algorithms and investigates the effectiveness of some bias mitigation algorithms in this problem domain. First, it collects data on various structural and neighborhood-level characteristics for 232,057 residential properties across San Antonio, TX. Then, it develops a set of commonly-used classical ML models and neural networks to predict the percentage of house price change over years, and studies the extent to which the outcome of such predictive models are fair with respect to the race and/or ethnicity. As a result, it finds that the models could be highly biased towards those protected attributes. Then, it investigates the effectiveness of CR and Reduction-based bias mitigation algorithms in reducing bias in this particular application. As a result, it finds that the effectiveness of those bias reduction methods could vary across different ML models, however, in general, the Reduction-based algorithm tends to be more effective than the CR in improving the fairness of such ML models.